\documentclass[10pt,twocolumn,letterpaper]{article}
\usepackage{iccv}
\definecolor{iccvblue}{rgb}{0.21,0.49,0.74}
\usepackage[pagebackref,breaklinks,colorlinks,allcolors=iccvblue]{hyperref}

\usepackage{amsmath}
\usepackage{amssymb}
\usepackage{booktabs}
\usepackage{multirow}
\usepackage{tabularx}
\usepackage{bm}
\usepackage{placeins}
\usepackage{afterpage}
\usepackage{enumitem}
\usepackage{wrapfig,lipsum,booktabs}
\usepackage{algorithm}
\usepackage{algpseudocode}
\usepackage{afterpage}
\usepackage{float}
\usepackage{colortbl}

\def\confName{ICCV}
\def\confYear{2025}

\title{Instruction-Grounded Visual Projectors for\\Continual Learning of Generative Vision-Language Models}

\author{
Hyundong Jin$^{1}$ ~~~~~~~~~~~~ Hyung Jin Chang$^{2}$ ~~~~~~~~~~~~  Eunwoo Kim$^{1,}\thanks{Corresponding author.}$ \\
$^{1}$School of Computer Science and Engineering, Chung-Ang University\\
$^{2}$School of Computer Science, University of Birmingham\\
{\tt\small \{jude0316, eunwoo\}@cau.ac.kr  ~~
        h.j.chang@bham.ac.uk}
}

\begin{document}
\maketitle
\begin{abstract}
Continual learning enables pre-trained generative vision-language models (VLMs) to incorporate knowledge from new tasks without retraining data from previous ones.
Recent methods update a visual projector to translate visual information for new tasks, connecting pre-trained vision encoders with large language models.
However, such adjustments may cause the models to prioritize visual inputs over language instructions, particularly learning tasks with repetitive types of textual instructions.
To address the neglect of language instructions, we propose a novel framework that grounds the translation of visual information on instructions for language models.
We introduce a mixture of visual projectors, each serving as a specialized visual-to-language translation expert based on the given instruction context to adapt to new tasks.
To avoid using experts for irrelevant instruction contexts, we propose an expert recommendation strategy that reuses experts for tasks similar to those previously learned.
Additionally, we introduce expert pruning to alleviate interference from the use of experts that cumulatively activated in previous tasks.
Extensive experiments on diverse vision-language tasks demonstrate that our method outperforms existing continual learning approaches by generating instruction-following responses.
\end{abstract}    
\section{Introduction}
Generative Vision-Language Models (VLMs) \cite{zhu2023minigpt, liu2024visual, lu2024unified, cao2024generative} have demonstrated remarkable performance in a variety of tasks, such as caption generation \cite{karpathy2015deep}, visual question answering \cite{ren2015exploring}, and creative applications \cite{zhu2023minigpt, achiam2023gpt}. 
Their impressive performance is attributable to the robust capabilities of large language models in conjunction with a translation module that enables these models to comprehend visual information from a visual encoder.
Despite their remarkable performance, generative VLMs face significant challenges in adapting to newly emerging tasks. 
Retraining these models with pre-trained and newly acquired data imposes considerable computational costs due to the large volume of pre-trained data \cite{lacoste2019quantifying} and its limited public accessibility \cite{radford2021learning}. 
Furthermore, training models solely on new tasks leads to overwriting previously acquired knowledge, resulting in catastrophic forgetting \cite{mccloskey1989catastrophic, kirkpatrick2017overcoming}.

\begin{figure}[t] 
    \centering 
    \includegraphics[width=\columnwidth]{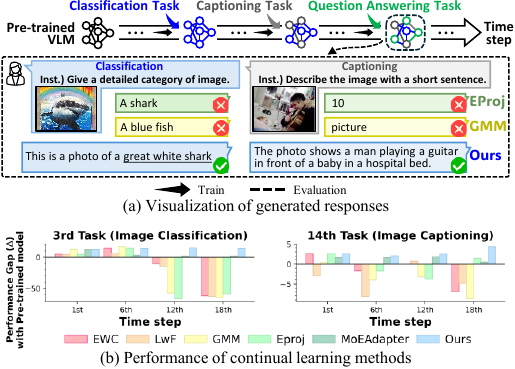}
    \caption{
    (a) Generated responses for classification and captioning tasks from continual learning methods for VLM after training on a question answering task.
    Existing methods, EProj \cite{he2023continual} and GMM \cite{cao2024generative}, neglect text instruction.
    (b) Performance gap between the pre-trained model and models trained on sequential tasks using various continual learning methods.
    The proposed method outperforms others in preventing knowledge forgetting.}
    \label{fig: intro overview} 
\end{figure}

Continual learning methods \cite{rebuffi2017icarl, kirkpatrick2017overcoming, liang2024inflora, wang2022learning, jin2022helpful, jin2023growing} aim to address these challenges by allowing models to incorporate new tasks while preserving old knowledge with access only to the data of newly emerging tasks.
Existing continual learning methods rely on storing a small set of previous data and training it jointly with the new data \cite{rebuffi2017icarl, douillard2022dytox} or on regularizing parameter updates that are important for previous tasks \cite{kirkpatrick2017overcoming}.
Other methods \cite{wang2022learning, smith2023coda, liang2024inflora} incorporate a small set of learnable parameters into pre-trained knowledge to capture task-specific knowledge.
Despite these advancements, continual learning for VLMs remains underexplored.
Recently, a few studies \cite{zhai2023investigating, cao2024generative} proposed to directly update the pre-trained visual projector of VLM that converts visual features for language models to learn new tasks, where tasks share similar instruction templates accompanied by similar answer types.
Although these methods show promising results for such tasks, they neglect text instructions for response generation after training on tasks with varying instructions, as shown in Figure \ref{fig: intro overview} (a).

In this paper, we propose a novel framework for transforming visual information grounded on language instructions for continual learning of VLMs. 
To this end, we introduce a \textbf{M}ixture of \textbf{V}isual \textbf{P}rojectors, named \textbf{MVP}, each serving as an expert dedicated to transforming visual information based on a given instruction context.
In particular, our routing scheme dynamically adapts to both the semantic content of the instruction and the visual information, leveraging translation experts that align with the linguistic intent.
To encourage the use of appropriate experts for given instructions, we propose an expert recommendation strategy that promotes reusing experts associated with relevant previous tasks while discouraging those specialized in other instructions.
In addition, we propose expert pruning to mitigate negative transfer to future tasks induced by the reuse of redundantly activated experts in previous tasks.
To preserve the zero-shot capabilities of pre-trained VLMs, we adaptively aggregate the transformed visual information from the proposed mixture of visual projectors and the pre-trained projector, calibrating their contributions based on the relevance of the given input data to the learned tasks.

To evaluate the proposed method, we perform vision-language tasks, including classification \cite{hendrycks2021many}, captioning \cite{plummer2017flickr30k}, and question answering \cite{ren2015exploring}, with a diverse range of visual and textual data. 
We evaluated its ability to retain new knowledge and maintain zero-shot performance, as illustrated in Figure \ref{fig: intro overview} (b).
The experimental results show that the proposed method outperforms its competitors by translating visual information for a given instructional context.
The contributions of our work are four-fold:
\begin{itemize}
\item We propose a novel framework for adaptively translating visual information grounded in textual instruction for continual learning of generative vision-language models.

\item We present expert recommendation and pruning strategies for sparse activation of experts from semantically relevant old tasks while mitigating negative knowledge transfer.

\item To preserve zero-shot capabilities, we adaptively aggregate knowledge by balancing pre-trained and instruction context-aware projections.

\item Extensive experiments show that the proposed framework outperforms other continual learning methods in preserving both pre-trained and newly learned knowledge.
\end{itemize}

\section{Related Work}
\noindent{\textbf{Generative Vision-Language Models}} (VLMs) \cite{li2023blip, dai2023instructblip, zhu2023minigpt, ye2023mplug, liu2024improved} demonstrate notable zero-shot generalization and in-context learning capabilities for vision-language tasks.
These models harness the robust capabilities of Large Language Models (LLMs) \cite{chiang2023vicuna, touvron2023llama} by employing connection modules that bridge the visual encoder with the LLM, e.g., Qformers \cite{li2023blip, dai2023instructblip} or resampler \cite{zhu2023minigpt, ye2023mplug, liu2024improved}.
By translating visual information into comprehensible representations for the language model, VLMs can interpret and generate responses based on complex visual and textual inputs.
Updating the connection module can directly improve the visual information for the language model with negligible parameter updates compared to updating the visual encoder and the language model \cite{cha2024honeybee, zhu2023minigpt}. 
However, this approach is vulnerable in that previously acquired information can be significantly compromised during adaptation to new vision-language tasks.
Therefore, a central challenge in VLMs is preserving the ability to follow previously learned instructions while adapting to newly introduced ones.

\noindent{\textbf{Continual Learning}} aims to learn from a stream of tasks without forgetting previously acquired knowledge \cite{mccloskey1989catastrophic, kirkpatrick2017overcoming}.
Existing continual learning methods incorporate additional regularization terms to penalize changes in critical parameters to maintain performance on earlier tasks \cite{aljundi2018memory, chaudhry2018riemannian, kirkpatrick2017overcoming, zenke2017continual}, or replay a memory buffer containing a small set of data from previous tasks when learning new tasks \cite{rebuffi2017icarl, douillard2022dytox, Liu2020AANets}.
Continual learning with parameter-efficient tuning methods \cite{wang2022learning, wang2022dualprompt, smith2023coda, gao2024beyond} learns task-specific representations by selectively updating a minimal subset of parameters \cite{hu2021lora, jia2022visual}.
Recently, several approaches for continual learning in visual language models \cite{cao2024generative, zhai2023investigating} incorporate a visual projector that is shared across tasks.
Although these approaches might be beneficial for tasks that share similar instruction templates, learning tasks that include contextually different instructions might cause the model to easily forget the previously learned visual representations.
In contrast, the proposed approach addresses this limitation by leveraging context-aware visual-to-language translation for given textual instructions.

\begin{figure*}[t] 
    \centering 
    \includegraphics[width=0.98\textwidth]{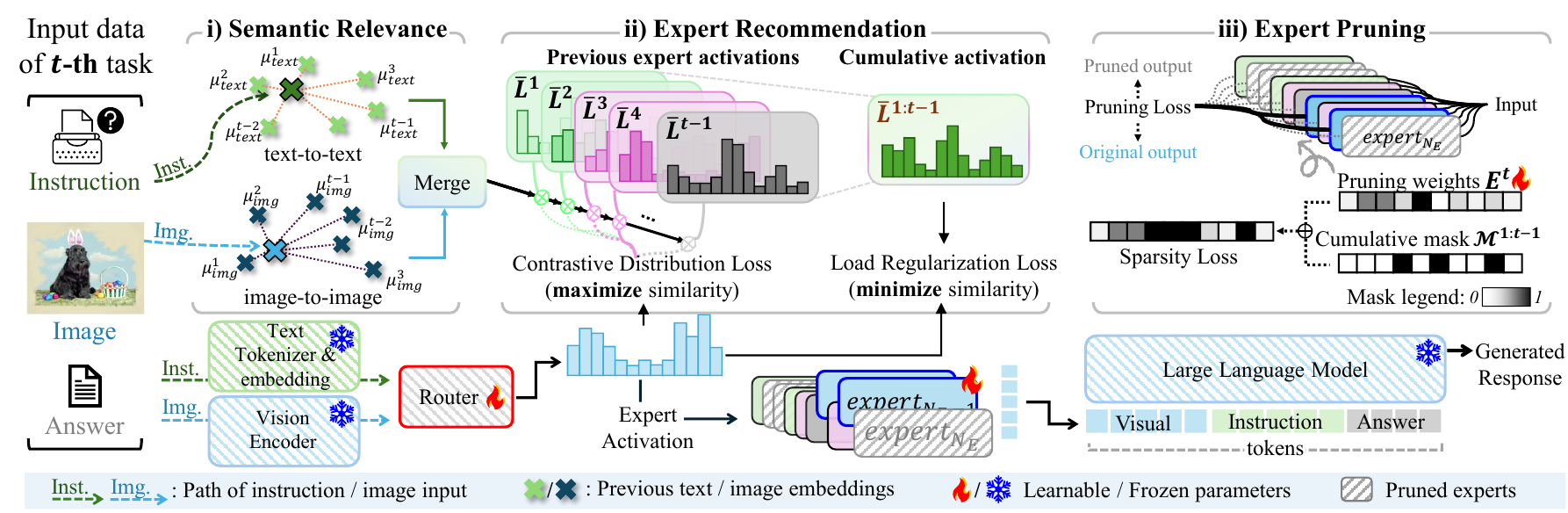}
    \caption{
     An illustration of the proposed continual learning framework for the $t$-th vision-language task.
     A router determines projector (expert) activations based on the embeddings of the instruction and image.
     To prevent detrimental expert activation of a router, \textbf{(\romannumeral 1)} we first evaluate the semantic relevance between the current and previous tasks within an embedding space.
    \textbf{(\romannumeral 2)} We then encourage the activation of experts used in semantically related previous tasks for new tasks, while suppressing those with high cumulative activation frequencies.
    \textbf{(\romannumeral 3)} To mitigate negative knowledge transfer to subsequent tasks, we prune and reinitialize experts with parameters of the pre-trained projector, thus securing learning capacity for subsequent tasks.
    Finally, the large language model generates responses through vision, instruction, and answer tokens.}
    \label{fig: overview} 
\end{figure*}

\noindent{\textbf{Mixture-of-Experts}} (MoE) \cite{jacobs1991adaptive, chen2023lifelong, lu-etal-2024-experts, mustafa2022multimodal, yu2024boosting} consists of multiple distinct networks, called experts, and a routing network to select appropriate experts and aggregate the outputs of the selected experts. 
In continual learning, MoE-based methods have been introduced \cite{aljundi2017expert, chen2023lifelong, yu2024boosting} by expanding the router \cite{aljundi2017expert, chen2023lifelong, yu2024boosting} or adding more experts \cite{aljundi2017expert, chen2023lifelong} in response to new tasks. 
Additionally, existing methods \cite{chen2023lifelong, yu2024boosting} prohibit updates to a subset of experts associated with old tasks to mitigate forgetting when learning new tasks.
However, this can potentially hinder learning for new tasks by freezing important experts, and a large number of frozen experts complicates capturing new representations.
In this work, we resolve these limitations by recommending experts to be updated based on their relevance to the new task while preventing reliance on irrelevant ones, ensuring that experts remain active to provide sufficient learning capacity for new tasks.
In addition, we propose expert pruning to break the cycle of negative knowledge transfer caused by regularizing balanced expert utilization, which risks subsequent tasks reusing redundantly updated experts from previous tasks.

\section{Methodology}
\subsection{Overview}
\noindent Continual learning aims to enable pre-trained generative vision-language models (VLMs) to preserve previously acquired knowledge while sequentially adapting to a sequence of novel tasks.
At time step $t$, VLM is trained on the $t$-th vision-lanugage task, where the $t$-th task $T^{t} = \{x^{t}_{i}, q^{t}_{i}, y^{t}_{i}\}_{i=1}^{N^{t}}$ comprises $N^{t}$ triplets, where $x^{t}_{i}$, $q^{t}_{i}$, and $y^{t}_{i}$ represent an image, a text instruction, and a text-based answer for the $i$-th sample, respectively.
For each triplet, the image feature $\bm{x}_{i,\text{img}}^{t} = \mathcal{F}_{v}(x^{t}_{i})$ is extracted by the vision encoder $\mathcal{F}_{v}$, and the instruction tokens $\bm{x}_{i,\text{text}}^{t} = \mathcal{F}_{e}(q^{t}_{i})$ and the answer tokens $\bm{y}_{i}^{t} = \mathcal{F}_{e}(y^{t}_{i})$ are extracted by the text embedding function $\mathcal{F}_{e}$ of the pre-trained model.

To make the image feature $\bm{x}_{i,\text{img}}^{t}$ compatible with the language model, a visual projector $\mathcal{V}$ transforms it into $\bar{\bm{x}}_{i, \text{img}}^{t} = \mathcal{V}(\bm{x}_{i,\text{img}}^{t})$.
Then the auto-regressive decoder $\mathcal{D}$ generates responses by receiving the tokens $\bar{\bm{x}}_{i, \text{img}}^{t}$, $\bm{x}_{i, \text{text}}^{t}$, and $\bm{y}_{i}^{t}$.
A na\"ive method to learn a new task is to update the pre-trained visual projector to align the predicted responses with the answer tokens by minimizing the cross-entropy loss as
$\mathcal{L}_{ce} = - \sum_{i=1}^{N^{t}} \sum_{j=1}^{L} \bm{y}_{i,j}^{t}  \log \hat{\bm{y}}_{i,j}^{t}, $
where $L$ denotes the number of tokens in the response.
$\bm{y}_{i, j}^{t}$ denotes the $j$-th token of the answer for the $i$-th sample, and $\hat{\bm{y}}_{i, j}^{t}$ denotes the corresponding predicted token.

Although updating the visual projector of a VLM is a parameter-efficient approach \cite{cao2024generative, he2023continual}, \textit{training it with similar repetitive instructions restricts its translation ability to reflecting only those instruction contexts.
This results in the model neglecting different text instructions}, as illustrated in Figure~\ref{fig: intro overview} (a).
To address this, we propose a novel framework that adaptively translates visual information via a mixture of projectors to provide instruction-context aware visual translation. 
To encourage the use of projectors for relevant contexts, we introduce a recommendation strategy that promotes the activation of projectors associated with semantically similar previous tasks.
Additionally, we present expert pruning to eliminate redundant experts that may cause negative transfer to subsequent tasks.
The overall framework is illustrated in Figure \ref{fig: overview}.

\subsection{Mixture-of-Visual Projectors}
We present a method to adaptively adjust visual projectors grounded on the text instructions of a new task.
Our framework includes a router $\mathcal{R}$ and a set of projectors (experts) $\{\mathcal{E}_{j}\}_{j=1}^{N_E}$, each contributing as a visual-to-language translator for the given instructional context.
For the $t$-th task, the router takes in multimodal data $\bm{x}_{i, \text{img}}^{t}$ and $\bm{x}_{i, \text{text}}^{t}$ to select the appropriate experts for translating $\bm{x}^{t}_{i,\text{img}}$ for a language model. 
The aggregated output from the selected experts is
\begin{equation}
    \tilde{\bm{x}}^{t}_{i,\text{img}} = \sum_{j=1}^{N_E} w_{j} \mathcal{E}_{j}(\bm{x}_{i, \text{img}}^{t}), 
\end{equation}
where $w_{j}$ denotes the routing weight determined by $\mathcal{R}$, which quantifies the contribution of the $j$-th expert.
To ensure the visual projector grounded on the text instructions of newer tasks, we define the weights $W=\{w_{j}\}_{j=1}^{N_E}$ to be conditioned on the instruction tokens as 
\begin{equation}
    W = \mbox{Softmax}(\mbox{Top-}K(\mathcal{R}(\bm{x}_{i, \text{img}}^{t}, \bm{x}_{i, \text{text}}^{t}))), 
\end{equation}
where $K$ denotes the pre-defined number of experts engaged for the translation.
To provide the language model with a richer representation, we integrate the translated token from the newly selected experts with the token obtained from the pre-trained projector $\mathcal{V}$ as 
\begin{equation}
    \tilde{\bm{x}}^{t}_{i,\text{img}} = \frac{1}{K+1}\left(\mathcal{V}(\bm{x}_{i, \text{img}}^{t}) + \sum_{j=1}^{N_E} w_{j} \mathcal{E}_{j}(\bm{x}_{i, \text{img}}^{t}) \right). 
\end{equation}

\subsubsection{Expert Recommendation}
To leverage the experts associated with old tasks that share similar visual-linguistic contexts with the new task, we introduce a recommendation strategy driven by the semantic relevance between old and new tasks.
Specifically, we compute similarity scores for vision and language instructions: $\bm{S}^{t}_{\text{img}} = \{s^{t'}_{\text{img}}\}_{t'=1}^{t-1}$, which measures the cosine similarity between $\bm{x}^{t}_{i, \text{img}}$ and the stored average image embedding $\mu^{t'}_{\text{img}}$ of the $t'$-th previous task. 
Similarly, the corresponding set of textual similarity scores, $\bm{S}^{t}_{\text{text}} = \{s^{t'}_{\text{text}}\}_{t'=1}^{t-1}$, is computed from the current input $x^{t}_{i, \text{text}}$ and the stored instruction embedding $\mu^{t'}_{\text{text}}$ \footnote{$\mu^{t'}_{\text{img}}= \text{Average}(\{x^{t'}_{i,\text{img}} \}_{i=1}^{N^{t'}})$ and $\mu^{t'}_{\text{text}}= \text{Average}(\{x^{t'}_{i,\text{text}}\}_{i=1}^{N^{t'}})$}. 
These are combined as $\bm{S}^{t} = \{s^{t'}\}_{t'=1}^{t-1}$, where 
\begin{equation}
\label{equation: score}
    s^{t'} = \alpha \cdot \sigma( s^{t'}_{\text{img}}) + (1-\alpha) \cdot \sigma(s^{t'}_{\text{text}}).
\end{equation}
$\sigma(\cdot)$ is the sigmoid function, and $\alpha$ balances the two scores. 
A higher score $s^{t'}$ indicates stronger alignment with the $t'$-th previous task.

Given these scores, we leverage a contrastive distribution loss \cite{tian2021constrained} to encourage the outputs of the router to associate activation patterns of similar tasks while differentiating them from dissimilar ones.
Let us denote the record of expert activation frequency from previous tasks by $\bar{L}^{t'}$ and the router output for the current task by ${L}^{t}$, both normalized via a Softmax function.
The recommendation loss becomes $\mathcal{L}_{rec} = \sum_{i=1}^{N^{t}}H(\bm{S}^{t}, \pi_{t,t'})$, where
\begin{equation}
    \pi_{t, t'} = \frac{\mbox{exp}(\mbox{sim}(\bar{L}^{t'}, L^{t}))}{\sum_{t'=1}^{t-1}\mbox{exp}(\mbox{sim}(\bar{L}^{t'}, L^{t})/\tau)}, 
\end{equation}
$H(\cdot)$ is the cross-entropy, $\mbox{sim}(\cdot)$ is the cosine similarity, and $\tau$ is the temperature parameter.
By emphasizing the expert patterns that correlate with previously encountered tasks, this recommendation encourages the router to reuse relevant expertise while minimizing reliance on less helpful ones.

\noindent{\textbf{Activation Bias Reduction.}} 
Although a mixture of experts provides flexibility in adapting to different tasks, they can lead the router to assign large weights to only a few experts \cite{shazeer2017outrageously, lepikhingshard}.
To mitigate such concentration, we introduce an additional regularization term that promotes diverse expert engagement by suppressing the activation of those specialized in previous tasks.
We minimize
\begin{equation}
    \mathcal{L}_{bias} = \frac{1}{2} \left(1 +  \frac{\langle \bar{L}^{1:t-1}, L^{t} \rangle}{\Vert \bar{L}^{1:t-1}\Vert_{2} \cdot \Vert L^{t}\Vert_{2}} \right), 
\end{equation}
where $\bar{L}^{1:t-1}$ is the normalized aggregated historical activation vector across all previous tasks.
$\langle \cdot, \cdot \rangle$ and $\Vert \cdot \Vert_{2}$ denote the inner product and the $\ell_2$ norm, respectively.
This promotes the activation of a diverse set of unused experts while aligning the activations of new tasks with those from previous related tasks.
To summarize, the proposed framework updates $\mathcal{R}$ and $\{\mathcal{E}_{j}\}_{j=1}^{N_{E}}$ by minimizing $\mathcal{L}=\mathcal{L}_{ce} + \lambda_{rec}\mathcal{L}_{rec} + \lambda_{bias} \mathcal{L}_{bias}$, where $\lambda_{rec}$ and $\lambda_{bias}$ are weighting factors for the respective losses.

\subsubsection{Expert Pruning}
A large number of previously activated experts may cause negative knowledge transfer to subsequent tasks.
To alleviate the interference, we propose expert pruning on a new task after training with $\mathcal{L}$.
We introduce a learnable vector, $E^{t}=\{e^{t}_{1},...,e^{t}_{N_E}\}$, which is updated to limit the number of experts used for both the new and previous tasks.
To discard redundant experts of the current task with sparsely activated experts, we minimize
\begin{equation}
    \min_{E^{t}} \left\Vert \sum_{j=1}^{N_E} (w_{j} - e^{t}_{j})\mathcal{E}_{j}(x^{t}_{i,\text{img}}) \right\Vert_{F} + \Vert \mathcal{M}^{1:t-1} + E^{t} \Vert_{1}, 
\end{equation}
where $\Vert \cdot \Vert_{F}$ and $\Vert \cdot \Vert_{1}$ denotes the Frobenius and the $\ell_{1}$ norm. 
By minimizing the discrepancy between the output of the original expert mixture and its pruned counterpart, while constraining activation to a few experts, we identify a sparse subset of experts for the current task.
The updated $E^{t}$ is thresholded to produce a binary mask $\mathcal{M}^{t}$.
After pruning the experts by the accumulated binary mask $\mathcal{M}^{1:t}$, the router is then fine-tuned to leverage the survived projectors for previous and new tasks.
To collect inputs for router fine-tuning, we sample from a normal distribution defined by the stored mean and covariance of images and text instructions from observed tasks $t'\in[1,t]$.
For each collected data, the corresponding label is sampled from $\mbox{Softmax}(L^{t'} \odot \mathcal{M}^{t'})$ over the indices of experts $1, 2, \ldots, N_E$.
We minimize the cross-entropy loss to fine-tune the router.

\subsection{Adaptive Knowledge Aggregation}
Preserving the generalization capability of pre-trained models developed on extensive datasets remains critical.
Our framework can effectively maintain this capability, as the proposed mixture of visual projectors is residually connected to the pre-trained projector.
To this end, we present an adaptive knowledge aggregation strategy in the inference phase by leveraging the contribution of the MoE branch based on the semantic relevance of the data to previously encountered information.

Specifically, after learning the $t$-th task, we have the stored average visual and instruction embeddings $\{ \mu^{t'}_{\text{img}}, \mu^{t'}_{\text{text}}\}_{t'=1}^{t}$.
The semantic relevance between the inference data and the stored embeddings is calculated as Eq.(\ref{equation: score}), yielding a combined set of scores $\bm{S}^{t}_{\text{inf}}=\{s^{t'}_{\text{inf}}\}_{t'=1}^{t}$.
Intuitively, a high relevance score $s^{t'}_{\text{inf}}$ indicates that the given vision-instruction paired data is more likely associated with the $t'$-th task.
Consequently, the translated visual token for the language model is represented as a linear combination of the pre-trained projection and the MoE-translated projection, weighted by $\lambda_{\text{inf}} = {\mathrm{argmax}}_{t'\in[1,t]} \{s^{t'}_{\text{inf}}\}_{t'=1}^{t}$, as
    \begin{equation}
         \frac{1}{\lambda_{\text{inf}}\cdot K+1}\left(\mathcal{V}(\bm{x}_{i, \text{img}}^{\text{inf}}) + \lambda_{\text{inf}} \sum_{j=1}^{N_E} w_{j} \mathcal{E}_{j}(\bm{x}_{i, \text{img}}^{\text{inf}}) \right), 
    \end{equation}
where $\bm{x}^{\text{inf}}_{i,\text{img}}$ denotes an input for inference and the weighting factor $\lambda_{\text{inf}}$ ranges in $[0,1]$. 
This adaptive aggregation ensures that the model retains its inherent zero-shot generalization capacity for unseen samples by calibrating the contributions of MoE-derived knowledge.
\section{Experiments}

\subsection{Experimental Setting}
\noindent\textbf{Datasets.}
We applied the proposed \textbf{M}ixture-of-\textbf{V}isual \textbf{P}rojectors, named \textbf{MVP}, to continual learning of diverse vision-language tasks; image classification \cite{hendrycks2021many}, image captioning \cite{plummer2017flickr30k}, and image question answering \cite{ren2015exploring}.
For classification, we employed the ImageNet-R dataset \cite{hendrycks2021many} widely used in continual learning \cite{wang2022dualprompt, smith2023coda}. 
The dataset was divided into 10 subsets containing 20 classes each, following the previous practices \cite{wang2022dualprompt, smith2023coda}.
We used Flickr-30K \cite{plummer2017flickr30k} for captioning tasks.
We divided the dataset into four subsets based on the type of visual categories \cite{del2020ratt}: animals, instruments, scenes, and vehicles.
For question answering, we employed the COCO-QA dataset \cite{ren2015exploring}.
We divided the dataset into four distinct subsets based on the type of language instructions \cite{zhang2024prompt}: object, count, color, and position.
We define each subset as a task and train one task at each time step, which results in a sequence of 18 tasks.

\noindent{\textbf{Evaluation metrics.}}
To evaluate the proposed method, we employ three metrics \cite{zheng2023preventing, yu2024boosting}: \textit{Last}, \textit{Transfer}, and \textit{Avg}.
\textit{Last} calculates the final performance after training all tasks. 
\textit{Transfer} measures the models' zero-shot performance on unseen tasks.
\textit{Avg} is the average performance across all time steps.
For classification and question answering tasks, we reported the metrics based on the accuracy of generated responses. 
For captioning tasks, the performance is evaluated by the RefCLIP-Score \cite{hessel2021clipscore}.
To enhance interpretability, we scaled the score, which ranges from 0 to 1.
Please refer to the supplementary materials for measuring the accuracy of the generated responses.

\noindent{\textbf{Implementation details.}}
We adopted Vicuna-7B \cite{chiang2023vicuna} as the language decoder and employed the pre-trained ViT-g/14 \cite{fang2023eva} in conjunction with the Qformer as the visual encoder, following the practice in \cite{cao2024generative}.
Additionally, we employed another VLM that utilized LLaMa-2-7B \cite{touvron2023llama} as the language decoder and the visual encoder \cite{fang2023eva} without Qformer.
We initialized the experts of the proposed method using the pre-trained visual projectors of VLMs, provided in \cite{zhu2023minigpt}.
We set the number of experts $N_E$ to 20, $K$ to 2, semantic score parameter $\alpha$ to 0.3, and the loss weights $\lambda_{rec}$ and $\lambda_{bias}$ to 1 for all experiments.
The proposed method was trained using the Adam optimizer with $\beta_1$ and $\beta_2$ of 0.9 and 0.999. 
All experiments were conducted using NVIDIA RTX 3090 GPUs.
For additional details, please refer to the supplementary material.

\noindent{\textbf{Comparison methods.}}
To evaluate the proposed method, we compared it with existing continual learning methods, including conventional methods, LwF \cite{li2017learning}, EWC \cite{kirkpatrick2017overcoming}, and MoEAdapter \cite{yu2024boosting}, and the methods for VLMs, GMM \cite{cao2024generative} and EProj \cite{he2023continual}.
All comparison methods employed the same pre-trained VLMs and froze the visual and language encoders while training the visual projector for tasks. 
For the MoE-based continual learning method \cite{yu2024boosting}, we utilized its routing scheme and set its expert as the visual projector.
Additionally, we reported the zero-shot performance of the pre-trained model, referred to as Zero-Shot.
\setlength{\fboxsep}{0.5pt}
\definecolor{lightblue}{RGB}{173,216,230}

\renewcommand{\arraystretch}{0.8}
\begin{table*}[t]
\centering
\caption{
Results of the compared methods that employed Vicuna were measured using the \colorbox{blue!10}{\textit{Last}}, \colorbox{red!10}{\textit{Avg}}, and \colorbox{green!10}{\textit{Transfer}} metrics.
The classification results were averaged across tasks.
The names of the captioning and question answering tasks are abbreviated based on their visual and instruction types, respectively.
}
\label{tab:main table}
\footnotesize
\resizebox{0.9\textwidth}{!}{%
\begin{tabular}{l|ccc|cccc|cccc}
\bottomrule
\multirow{2}{*}{Method} \hspace{1em} & \multicolumn{3}{c|}{Classification (Accuracy)} & \multicolumn{4}{c|}{Captioning (RefCLIP-Score)}   & \multicolumn{4}{c}{Question Answering (Accuracy)}  \\ \cline{2-12}
                        & $T^{1}$-$T^{3}$ & $T^{4}$-$T^{6}$ & $T^{7}$-$T^{10}$  
                        & Anim. & Inst. & Scen. & Vehi.
                        & Obje. & Coun. & Colo. & Posi. \\ \hline \hline
\textbf{Vicuna}
&       &       &                            
&       &       &        &          
&       &       &        &            \\ 
\hspace{1em}Zero-Shot               
& 67.79 & 63.16 & 62.87 
& 77.08 & 75.24 & 74.24 & 73.43
& 68.52 & 42.55 & 62.62 & 32.79           \\ \hline \hline
\colorbox{blue!10}{\textit{\textbf{Last}}}
& & & & & & & & & & &            \\
\hspace{1em}LwF \cite{li2017learning}
& 4.38 & 4.77 & 5.83 
& 70.16 & 70.31 & 70.15 & 68.60
& 70.69 & 65.45 & 76.46 & 59.51           \\
\hspace{1em}EWC \cite{kirkpatrick2017overcoming}
& 5.61 & 6.80  & 5.83 
& 67.00 & 67.39 & 66.13 & 66.47
& 65.72 & 63.63 & 73.69 & 44.13         \\
\hspace{1em}GMM \cite{cao2024generative}
& 1.33 & 0.94 & 1.62 
& 62.96 & 61.08 & 63.73 & 64.70
& 42.29 & 46.23 & 45.11 & 25.74          \\
\hspace{1em}EProj \cite{he2023continual}
& 8.50 & 12.82 & 11.81 
& 73.78 & 71.43 & 72.39 & 72.23
& 39.77 & 60.00 & 65.84 & 24.69           \\
\hspace{1em}MoEAdapter \cite{yu2024boosting} 
& 70.01 & 65.10 & 65.55
& 76.53 & 76.23 & 76.43 & 73.92 
& 66.76 & 42.18 & 41.53 & 38.86          \\
\rowcolor[gray]{0.93}
\hspace{1em}MVP (Ours)
& \textbf{82.81} & \textbf{85.34} & \textbf{87.52}
& \textbf{79.60}  & \textbf{77.57}  & \textbf{76.06} & \textbf{77.79} 
& \textbf{79.26} & \textbf{74.54} & \textbf{80.30} & \textbf{71.25}          \\  \hline \hline
\colorbox{red!10}{\textit{\textbf{Avg}}}                
& & & & & & & & & & &            \\
\hspace{1em}LwF \cite{li2017learning}
& 48.37  & 49.42 & 51.62 
& 70.63  & 70.34 & 68.56 & 69.18
& 53.99 & 47.21 & 60.22  & 37.51          \\
\hspace{1em}EWC \cite{kirkpatrick2017overcoming}
& 55.21 & 50.13 & 52.74 
& 72.18 & 72.72 & 71.86 & 71.64
& 52.56 & 48.74 & 56.45 & 27.08          \\
\hspace{1em}GMM \cite{cao2024generative}
& 43.92 & 42.08 & 50.70
& 69.62 & 69.15 & 68.84 & 68.80 
& 48.39 & 48.18 & 48.47 & 27.71          \\
\hspace{1em}EProj \cite{he2023continual}
& 49.18 & 45.55 & 48.23
& 73.78 & 71.43 & 72.39 & 72.23
& 40.06 & \textbf{50.88} & 50.65 & 21.68          \\
\hspace{1em}MoEAdapter \cite{yu2024boosting}
& 71.50 & 67.16 & 66.27
& 76.65 & 75.97 & 74.40 & 74.66
& 67.95 & 42.60 & 53.82 & 36.48           \\
\rowcolor[gray]{0.93}
\hspace{1em}MVP (Ours)
& \textbf{81.34} & \textbf{82.03} & \textbf{85.07} 
& \textbf{78.71} & \textbf{76.92} & \textbf{75.92} & \textbf{76.25}
& \textbf{70.16} & 50.68 & \textbf{71.34} & \textbf{38.55}          \\ \hline \hline
\colorbox{green!10}{\textit{\textbf{Transfer}}}
&  &  &  
&  &  &  & 
&  &  &  &           \\
\hspace{1em}LwF \cite{li2017learning}
& 69.68 & 64.30 & 71.26 
& 67.15 & 67.97 & 66.74 & 68.41 
& 49.07 & 43.80 & 58.47 & 36.17          \\
\hspace{1em}EWC \cite{kirkpatrick2017overcoming}
& 71.85 & 66.81 & 75.69 
& 74.00 & 72.85 & 72.65 & 72.63
& 48.57 & 45.33 & 54.20 & 26.07           \\
\hspace{1em}GMM \cite{cao2024generative}
& 74.59 & 65.82 & 81.96 
& 70.40 & 70.30 & 69.84 & 69.77
& 42.29 & 46.23 & 45.11 & 25.74          \\
\hspace{1em}EProj \cite{he2023continual}
& 71.85 & 66.81 & 75.76
& 73.98 & 72.34 & 72.06  & 71.86 
& 42.17 & \textbf{48.99} & 48.65 & 21.50           \\
\hspace{1em}MoEAdapter \cite{yu2024boosting}
& 75.01 & 66.61 & 67.05
& 76.80 & 75.80 & 74.38 & 74.94 
& 68.29 & 42.69 & 55.35 & 36.34          \\
\rowcolor[gray]{0.93}
\hspace{1em}MVP (Ours)
& \textbf{77.95} & \textbf{73.34} & \textbf{82.52} 
& \textbf{78.02} & \textbf{76.88} & \textbf{75.71} & \textbf{75.91}
& \textbf{67.52} & 45.89 & \textbf{70.17} & \textbf{36.62}          \\ \toprule

\end{tabular}%
}
\renewcommand{\arraystretch}{1.0}
\end{table*}

\begin{figure*}[h] 
    \centering 
    \includegraphics[width=0.95\textwidth]{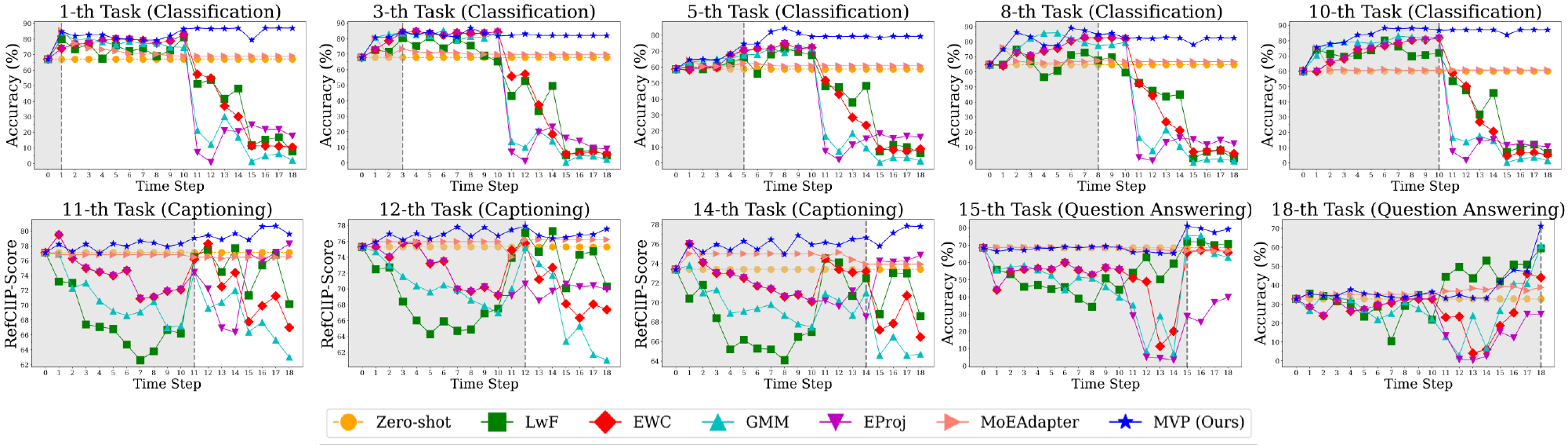}
    \caption{Performance of each task across time steps.  
    The gray area represents the time steps preceding training for specific tasks, and the vertical dashed line indicates the time step at which the corresponding task is learned.
    Best viewed in color ($\times$2).}
    \label{fig: main results} 
\end{figure*}

\subsection{Comparison Results}
We evaluated the proposed method and the comparison methods on diverse vision-language tasks learned in the order of classification, captioning, and question answering.
Table \ref{tab:main table} shows the experimental results of continual learning methods using Vicuna as a large language model.
Overall, the proposed method outperforms other continual learning methods in terms of the \textit{Last}, \textit{Avg}, and \textit{Transfer} metrics across most tasks.
The existing methods, LwF, EWC, GMM, and EProj, show unsatisfactory performance measured with \textit{Last} due to overwriting previously learned knowledge in a shared projector across tasks by the latest learned question answering task.
The proposed method, MVP, achieves superior classification performance, with notable performance gaps ranging from 12.80\% to 75.71\% compared to EProj and MoEAdapter. 
These results highlight the importance of instruction context-aware translation in continual learning for VLMs.

Note that most compared methods using a shared visual projector, which tends to overfit on classification, achieve higher performance than Zero-Shot in \textit{Transfer} but exhibit a significant performance drop in captioning and question answering tasks.
In contrast, MVP achieves higher performance than Zero-Shot in most tasks, demonstrating that the relevance-based expert recommendation and knowledge aggregation contribute to effective context-aware translation.
For the results of a different task order, please refer to the supplementary material.

We also present the performance of tasks at each time in Figure \ref{fig: main results}.
As shown in the figure (top), most methods show improved performance on subsequent classification tasks after training on the other classification tasks.
However, their performance significantly degrades after training on captioning and question answering tasks, with results lower than the zero-shot performance of the pre-trained model (indicated by the yellow horizontal line).
The performance of captioning tasks decreases when newer tasks are introduced, with this decline becoming more pronounced following the introduction of the question answering tasks.
In contrast, the proposed approach does not exhibit a performance drop in each task after learning other tasks.
Notably, the performance of captioning tasks shows a slight improvement after learning classification tasks.
This is attributed to the instruction-grounded translator that leverages knowledge of classification to help perform captioning tasks.

\newpage
To evaluate the proposed method based on another VLM, we conducted additional experiments using LLaMa-2 \cite{touvron2023llama}, which has a different language decoder and a vision encoder compared to the previous VLM using Vicuna.
Table \ref{tab:another VLM} shows the results of the methods with the same task order as in Table \ref{tab:main table}.
The trends of the results are similar to those in Table \ref{tab:main table}.
Most competitors underperform Zero-Shot in \textit{Last}, whereas MVP outperforms Zero-Shot by large average margins of 20.78\%, 2.82, and 7.15\% in classification, captioning, and question answering, respectively.
MVP also outperforms the runner-up method, MoEAdapter, in \textit{Transfer} by 9.24\%, 2.59, and 0.89\% across the tasks.
This results from relevance-based calibration of newly acquired and pre-trained knowledge, unlike MoEAdapter, which relies solely on pre-trained knowledge for unseen tasks.


\renewcommand{\arraystretch}{0.85}
\begin{table}[t]
\centering
\caption{
Results of using VLM with LLaMa-2. 
$\Delta$ indicates the performance gap with the pre-trained VLM. Average indicates the average performance of tasks.}
\label{tab:another VLM}
\huge
\resizebox{\columnwidth}{!}{%
\begin{tabular}{l|cc|cc|cc}
\bottomrule
\multirow{2}{*}{Method} \hspace{1em} & \multicolumn{2}{c|}{Classification} & \multicolumn{2}{c|}{Captioning}   & \multicolumn{2}{c}{\begin{tabular}[c]{@{}c@{}}Question\\ Answering\end{tabular}}  \\ \cline{2-7}
                        & Average & $\Delta~(\uparrow)$
                        & Average & $\Delta~(\uparrow)$
                        & Average & $\Delta~(\uparrow)$ \\ \hline \hline
\textbf{LLaMa-2}
&       &                            
&       &                 
&       &                   \\
\hspace{1em} Zero-Shot               
& 59.21 & 0.00 
& 75.90  & 0.00
& 60.31 & 0.00 \\ \hline \hline
\colorbox{blue!10}{\textit{\textbf{Last}}}       
& & & & & &           \\
\hspace{1em}LwF \cite{li2017learning}
& 70.13 & +10.92
& 77.12 & +1.22
& 54.79 & -5.52          \\
\hspace{1em}EWC \cite{kirkpatrick2017overcoming}
& 52.55 & -6.66
& 68.87 & -7.03
& 31.60 & -28.71          \\
\hspace{1em}GMM \cite{cao2024generative}
& 21.28 & -37.93
& 78.92 & +3.02
& 59.48 & -0.83          \\
\hspace{1em}EProj \cite{he2023continual}
& 53.68 & -5.53
& 66.42 & -9.48
& 30.44 & -29.87          \\
\hspace{1em}MoEAdapter \cite{yu2024boosting} 
& 61.30 & +2.09
& 75.87 & -0.03
& 59.26 & -1.05          \\
\rowcolor[gray]{0.93}
\hspace{1em}MVP (Ours)
& \textbf{79.99} & \textbf{+20.78}
& \textbf{78.72} & \textbf{+2.82}
& \textbf{67.46} & \textbf{+7.15}   \\  
 \hline \hline
\colorbox{red!10}{\textit{\textbf{Avg}}}    
& & & & & &           \\
\hspace{1em}LwF \cite{li2017learning}
& 68.02 & +8.81
& 73.50 & -2.40
& 49.92  & -10.39 \\
\hspace{1em}EWC \cite{kirkpatrick2017overcoming}
& 48.91 & -10.30
& 65.57 & -10.33
& 35.31 & -25.00        \\
\hspace{1em}GMM \cite{cao2024generative}
& 64.20 & +4.99
& 73.31 & -2.59
& 50.13 & -10.18          \\
\hspace{1em}EProj \cite{he2023continual}
& 49.74 & -9.47
& 64.61 & -11.29
& 32.82 & -27.49         \\
\hspace{1em}MoEAdapter \cite{yu2024boosting} 
& 62.77 & +3.56
& 75.87 & -0.03
& 60.16 & -0.15           \\
\rowcolor[gray]{0.93}
\hspace{1em}MVP (Ours)
& \textbf{78.58} & \textbf{+19.37}
& \textbf{78.37} & \textbf{+2.47}
& \textbf{61.66} & \textbf{+1.35}          \\ \toprule \bottomrule
\colorbox{green!10}{\textbf{\textit{Transfer}}}
&  &  
&  &  
&  &            \\
\hspace{1em}LwF \cite{li2017learning}
& 63.88  & +4.67
& 71.81  & -4.09
& 49.07  &  -11.24     \\
\hspace{1em}EWC \cite{kirkpatrick2017overcoming}
& 47.33 & -11.88 
& 64.24 & -11.66 
& 37.36 & -22.95     \\
\hspace{1em}GMM \cite{cao2024generative}
& 49.74  & -9.47
& 64.61  & -11.29
& 32.82  & -27.49         \\
\hspace{1em}EProj \cite{he2023continual}
& 47.11 & -12.10
& 63.92  & -11.98
& 34.19 &  -26.12         \\
\hspace{1em}MoEAdapter \cite{yu2024boosting} 
& 65.15 & +5.94
& 75.08 & -0.82 
& 56.27 & -4.04          \\
\rowcolor[gray]{0.93}
\hspace{1em}MVP (Ours)
& \textbf{74.39} & \textbf{+15.18}
& \textbf{77.67} & \textbf{+1.77}
& \textbf{57.16} & \textbf{-3.15}        \\ \toprule
\end{tabular}%
}
\renewcommand{\arraystretch}{1.0}
\end{table}

\subsection{Ablation Study and Analysis}
\noindent{\textbf{Ablation study.}}
We conducted an ablation study by sequentially removing key components: visual projectors, recommendation loss $\mathcal{L}_{rec}$, activation bias reduction loss $\mathcal{L}_{bias}$, expert pruning, and adaptive knowledge aggregation (AKA). 
Table \ref{tab:ablation study} shows the results of the ablation study.
The method without AKA degrades not only \textit{Transfer} but also \textit{Last} compared to that with AKA, demonstrating the importance of calibrating learned knowledge based on semantic relevance. 
The method without pruning degrades the average performance of captioning (76.70 to 71.55) and question answering (76.17\% to 75.54\%) in \textit{Last}, indicating that pruning helps to mitigate negative transfer by eliminating redundant experts.
We also observe that the method without AKA and pruning shows slightly lower performance in \textit{Transfer} than the approach that also removes $\mathcal{L}_{bias}$, as a diverse set of activated experts for seen tasks is repurposed for unseen tasks.
However, the methods using $\mathcal{L}_{bias}$ yield substantial improvement compared to the method without it in \textit{Last}, especially in classification (3.31\% to 85.92\%), by promoting diversified experts utilization.
Additionally, the methods without $\mathcal{L}_{rec}$ shows performance drop measured in \textit{Avg}, particularly in classification ($\Delta$ +5.18 to -5.90) and question answering ($\Delta$ +0.06 to -0.23).

\begin{figure*}[t] 
    \centering 
    \includegraphics[width=\textwidth]{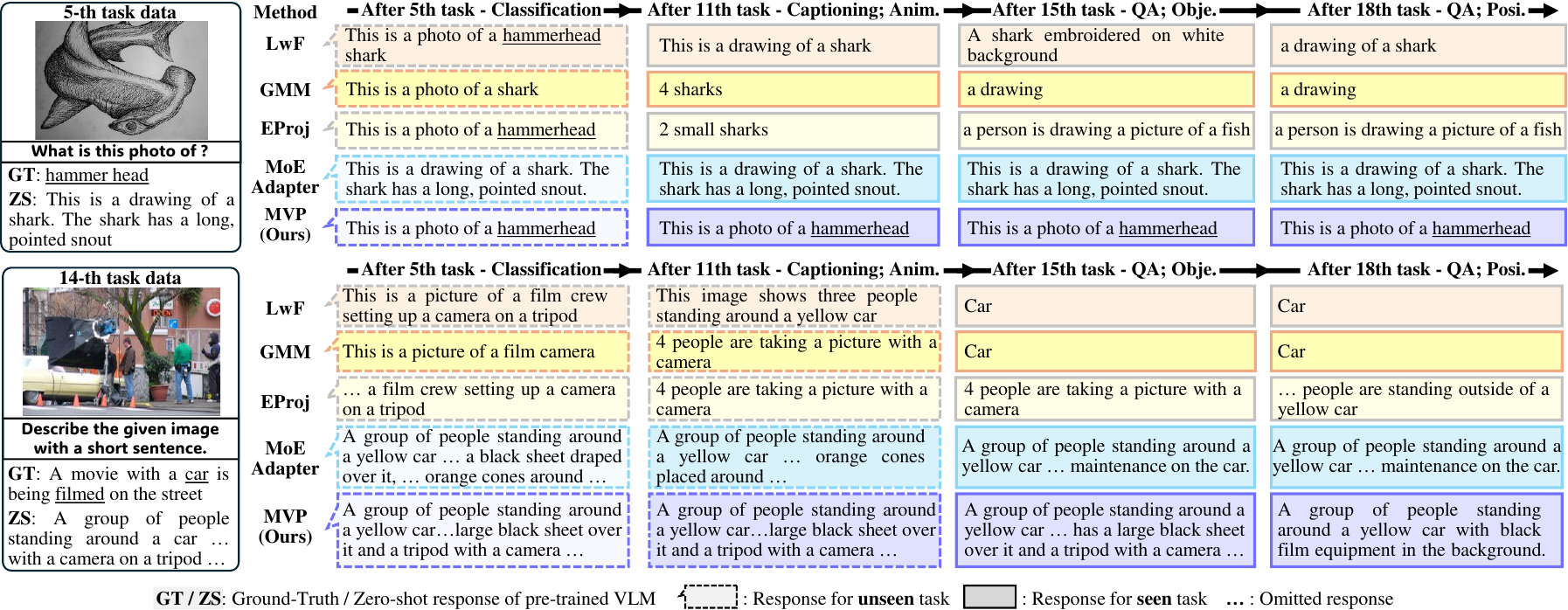}
    \caption{Qualitative results of the continual learning methods illustrating changes of responses to randomly selected samples after learning of newly emerging tasks.
    The proposed method generates consistent responses after learning other tasks, preserving both detailed category knowledge (top) and overall scene perception (bottom).
    }
    \label{fig:generated responses} 
\end{figure*}

\noindent\textbf{Qualitative results.}
To investigate the changes in responses throughout the continual learning process, we provide a qualitative analysis.
Figure \ref{fig:generated responses} shows generated responses over time steps from different continual learning methods using Vicuna.
The proposed method generates responses that reflect the given instruction for the visual input while maintaining the response quality similar to zero-shot performance, even before learning the task associated with the given image-instruction pair. 
In contrast, LwF and GMM often neglect the language instruction and forget previously learned detailed visual information, generating coarse-grained responses (e.g., ``a drawing'' for classification and ``car'' for captioning as shown in Figure \ref{fig:generated responses}).
By leveraging the instruction-aware visual projectors, MVP generates specialized responses for learned tasks and responses similar to zero-shot performance for unseen tasks.

\renewcommand{\arraystretch}{0.9}
\begin{table}[t]
\caption{Ablation study of MVP with different components.}
\centering
\label{tab:ablation study}
\huge
\resizebox{\columnwidth}{!}{%
\begin{tabular}{l|cc|cc|cc}
\bottomrule
\multirow{2}{*}{Method}  & \multicolumn{2}{c|}{Classification} & \multicolumn{2}{c|}{Captioning}   & \multicolumn{2}{c}{\begin{tabular}[c]{@{}c@{}}Question\\ Answering\end{tabular}}  \\ \cline{2-7}
                        & Average & $\Delta~(\uparrow)$
                        & Average & $\Delta~(\uparrow)$
                        & Average & $\Delta~(\uparrow)$ \\ \hline \hline
\textbf{Vicuna}
&       &                            
&       &                 
&       &                   \\
\hspace{0.5em}Zero-shot 
& 64.41 & 0.00 
& 75.00 & 0.00  
& 51.62 & 0.00      \\ \hline \hline
\colorbox{blue!10}{\textit{\textbf{Last}}}     
& & & & & &           \\
\hspace{0.5em}MVP (Complete method)
& 85.87 & +21.46
& 77.75 & +2.75
& 76.34 & +24.72           \\
\hspace{0.5em}w/o AKA
& 85.53 & +21.12
& 76.70 & +1.70
& 76.17 & +24.55          \\
\hspace{0.5em}w/o AKA, Prune
& 85.92 & +21.51
& 71.55 & -3.45
& 75.54 & +23.92     \\
\hspace{0.5em}w/o AKA, Prune, $\mathcal{L}_{bias}$
& 3.31  & -61.10 
& 70.70 & -4.30 
& 73.99 & +22.37           \\
\hspace{0.5em}w/o AKA, Prune, $\mathcal{L}_{bias}$, $\mathcal{L}_{rec}$
& 0.25  & -64.16
& 68.51 & -6.49
& 74.53 & +22.91      \\
\hspace{0.5em}w/o all (Single projector \cite{cao2024generative})
& 1.49 &  -62.92
& 63.10 & -11.90 
& 60.15 & +8.53           \\
 \hline \hline
\colorbox{red!10}{\textit{\textbf{Avg}}}       
& & & & & &           \\
\hspace{0.5em}MVP (Complete method)
& 83.28 & +18.87
& 76.94 & +1.94
& 57.68 & +6.06       \\ 
\hspace{0.5em}w/o AKA
& 83.02 & +18.61
& 72.21 & -2.79
& 48.64 & -2.98         \\
\hspace{0.5em}w/o AKA, Prune
& 83.51 & +19.10
& 69.22 & -5.78
& 47.05 & -4.57     \\
\hspace{0.5em}w/o AKA, Prune, $\mathcal{L}_{bias}$
& 69.59 & +5.18
& 71.60 & -3.40
& 51.68 & +0.06 \\
\hspace{0.5em}w/o AKA, Prune, $\mathcal{L}_{bias}$, $\mathcal{L}_{rec}$
& 58.51 & -5.90
& 70.59 & -4.41 
& 51.39 & -0.23    \\
\hspace{0.5em}w/o all (Single projector \cite{cao2024generative})
& 46.97 & -17.44 
& 69.10 & -5.90
& 43.10 & -8.52     \\
\hline \hline
\colorbox{green!10}{\textit{\textbf{Transfer}}}
&  &  
&  &  
&  &            \\
\hspace{0.5em}MVP (Complete method)
& 78.45 & +14.04
& 76.16 & +1.16
& 50.89 & -0.73    \\
\hspace{0.5em}w/o AKA
& 78.40 & +13.99
& 68.92 & -6.08
& 44.55 & -7.07    \\
\hspace{0.5em}w/o AKA Prune
& 78.34 & +13.93
& 68.78 & -6.22
& 43.03 & -8.59     \\
\hspace{0.5em}w/o AKA, Prune, $\mathcal{L}_{bias}$
& 78.06 & +13.65
& 71.47 & -3.53
& 47.94 & -3.68     \\
\hspace{0.5em}w/o AKA, Prune, $\mathcal{L}_{bias}$, $\mathcal{L}_{rec}$
& 78.26 & +13.85
& 70.45 & -4.55
& 47.68 & -3.94    \\
\hspace{0.5em}w/o all (Single projector \cite{cao2024generative})
& 76.31 & +11.90
& 69.97 & -5.03
& 39.03 & -12.59           \\
 \toprule
\end{tabular}%
}
\end{table}

\begin{figure}[t] 
    \centering 
    \includegraphics[width=\columnwidth]{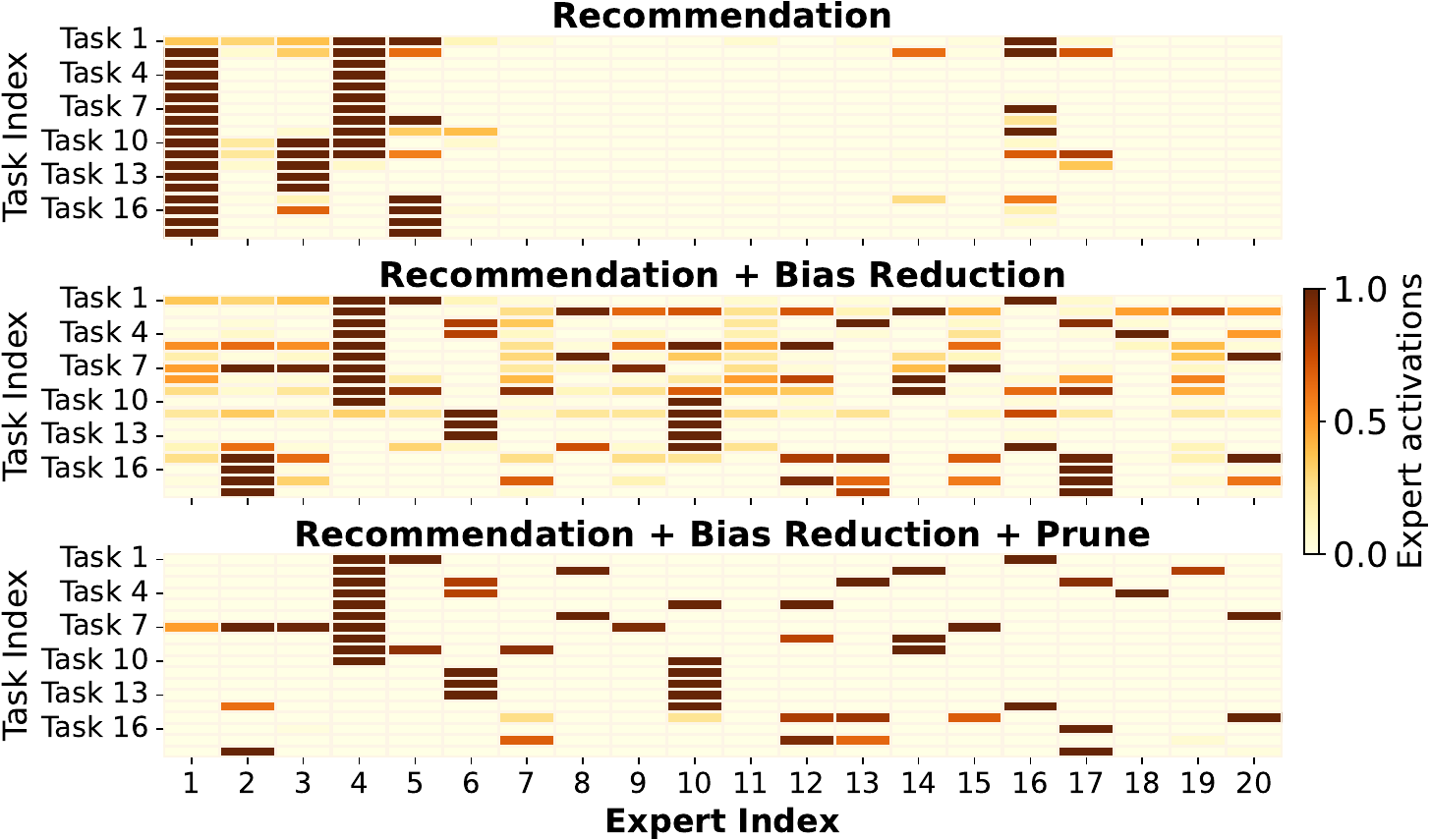}
    \caption{Results of expert activation frequency for each task by ablating the components of MVP.}
    \label{fig: expert activations} 
\end{figure}

\noindent{\textbf{Expert activation frequency.}}
We further analyze the combinations of three components, recommendation loss, activation bias reduction loss, and expert pruning, that influence the expert activation pattern, as in Figure \ref{fig: expert activations}.
When only the recommendation loss is applied, a limited subset of experts is predominantly activated (similar observations in \cite{shazeer2017outrageously, lepikhingshard}).
In contrast, combining recommendation with the activation bias reduction loss promotes broader expert activation. 
Notably, expert activation frequencies change at task transitions, particularly from classification to captioning (task index 11) and from captioning to question answering (task index 15).

\begin{figure}[t] 
    \centering 
    \includegraphics[width=\columnwidth]{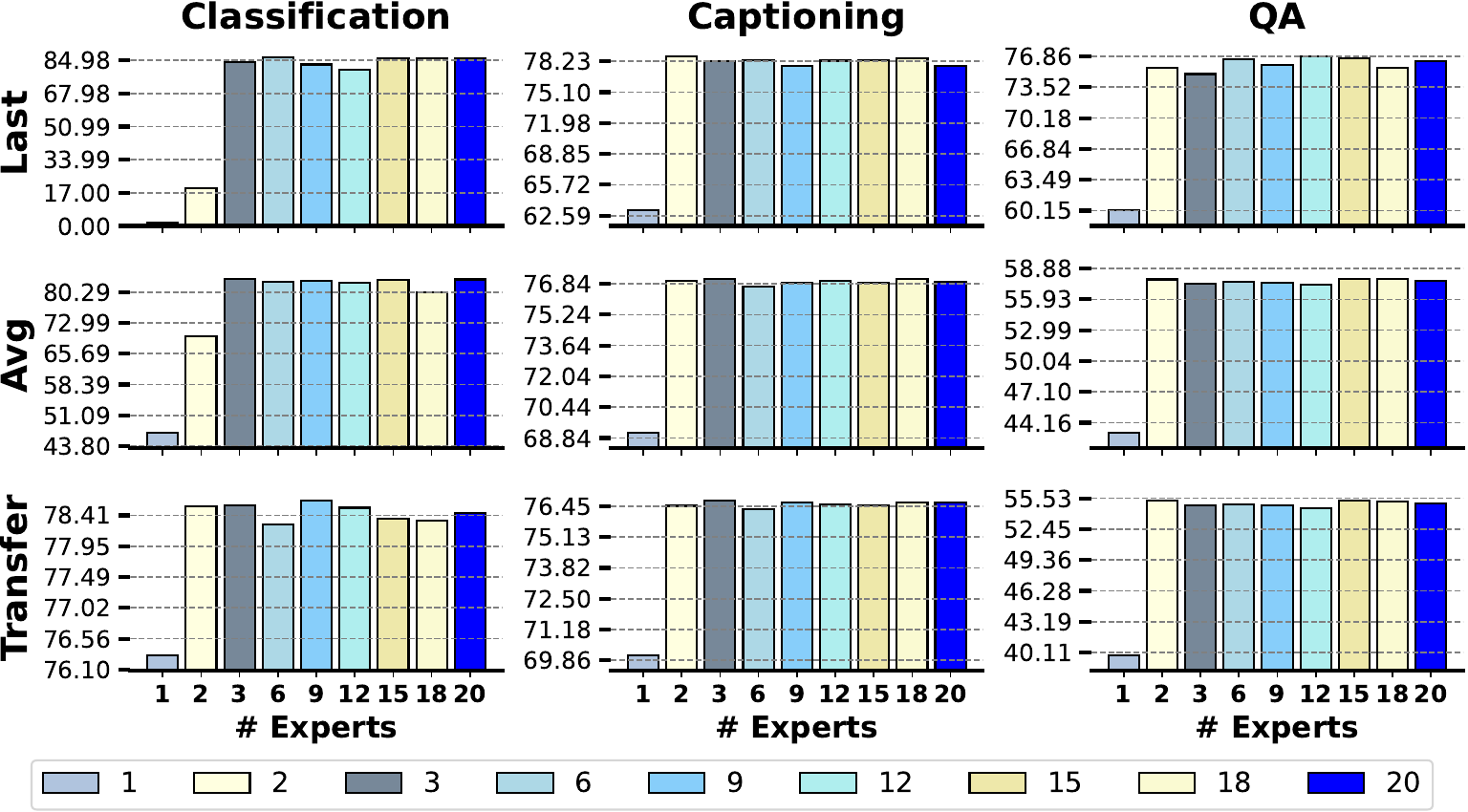}
    \caption{Results for MVP with varying numbers of experts.}
    \label{fig: albation n experts} 
\end{figure}

\noindent{\textbf{Impact of the number of experts.}}
We conducted additional experiments by varying the number of visual projectors in the proposed method.
The results presented in Figure \ref{fig: albation n experts} indicate that using a small number of experts (one or two) leads to a considerable degradation in performance. 
In contrast, incorporating three or more experts yields performance improvements and suggests that the additional experts provide sufficient capacity for tasks with novel language instructions.
Please refer to the supplementary material for additional results on varying text instructions, computational cost, and generative VLM benchmark \cite{li2023seed}.

\section{Conclusion}
In this paper, we have proposed a novel continual learning framework for generative vision-language models that adaptively translate visual information grounded on language instructions. 
We have introduced a mixture of visual projectors to leverage an appropriate visual translator based on the context of the given instruction and visual information. 
To mitigate the use of projectors across different contexts, we have proposed a semantic relevance-guided expert recommendation strategy that promotes the sparse activation of experts relevant to previous tasks while suppressing negative knowledge transfer.
Extensive experiments on a challenging sequence of tasks, including image classification, captioning, and question answering, demonstrate that the proposed method outperforms other continual learning methods on both seen and unseen tasks. 

{\noindent{\textbf{Acknowledgement.}}} 
This work was supported in part by the National Research Foundation of Korea (NRF) grant funded by the Korea government (MSIT) (RS-2023-00279019) and in part by the Institute of Information \& Communications Technology Planning \& Evaluation (IITP) grant funded by the Korea government (MSIT) (RS-2021-II211341, Artificial Intelligence Graduate School Program (Chung-Ang University)).

{
    \small
    \bibliographystyle{ieeenat_fullname}
    \bibliography{main}
}
\clearpage
\setcounter{page}{1}
\maketitlesupplementary

\renewcommand{\thetable}{\Alph{table}}
\renewcommand{\thefigure}{\Alph{figure}}
\renewcommand{\thesection}{\Alph{section}}
\setcounter{table}{0}
\setcounter{figure}{0}
\setcounter{section}{0}

\section{Additional Implementation Details}
\noindent\textbf{Details of Router.} 
We concatenated the image feature and instruction token and used them as input to the router.
To align the dimension of the image feature with that of the instruction token, we first applied a pre-trained visual projector to the image feature.
To more effectively extract contextual information from the given instructions, we employed a simple linguistic preprocessing technique, following \cite{conti2023vocabulary}. 
Specifically, we applied Part-of-Speech (POS) tagging to categorize key components, such as WH-words, nouns, and verbs, while filtering out irrelevant terms.
The filtered instructions were then processed through the tokenizer and embedding function of a pre-trained model to extract text tokens for the router. 
Note that this process was not applied to the instruction tokens for the pre-trained large language model. 
Finally, the projected image feature was concatenated with the filtered instruction token and used as input to the router.

To activate an expert by considering both the given image and instruction, we implemented the router consisting of a single multi-head self-attention block followed by a linear layer.
We set the number of translation experts $K$ engaged in translation to two for all experiments, following previous practices \cite{lu-etal-2024-experts, yu2024boosting}.
We pruned experts by applying a threshold to $E^{t}$, removing experts whose corresponding values in $E^{t}$ were smaller than $10^{-3}$.
After identifying redundant experts, we reinitialized those that had not been activated across all tasks learned thus far (i.e., zero-valued entries in the cumulative activation vector $\mathcal{M}^{1:t}$) using the parameters of the visual projector from the pre-trained model.

\noindent\textbf{Details of the Language Instructions.}
The language instructions used as the default setting in our experiments are presented in Table \ref{tab: details of instruction}.
For question answering tasks, the instruction varies depending on the sample.

\noindent\textbf{Measuring Accuracy of Generated Responses.}
For the classification and question answering tasks, the performance of the generated responses is evaluated in terms of accuracy. 
To measure accuracy for classification tasks, we compute the similarity in the embedding space between the textual representations of the classification categories and the generated responses.
The predicted class is the category with the highest similarity to the generated response; if it matches the ground truth, the response is considered correct.
To compute textual similarity, we extract embeddings using the text encoder of CLIP \cite{radford2021learning}, following \cite{cao2024generative}.
For question answering tasks, the accuracy is determined by whether the answer text appears in the generated response, following previous practices \cite{liu2024visual, chen2025coin}.

\renewcommand{\arraystretch}{0.85}
\begin{table}[t]
\caption{{Results on a different task order.}}
\centering
\label{tab:different task order}
\huge
\resizebox{\columnwidth}{!}{%
\begin{tabular}{l|cc|cc|cc}
\bottomrule
\multirow{2}{*}{Method} \hspace{1em} & \multicolumn{2}{c|}{Classification} & \multicolumn{2}{c|}{Captioning}   & \multicolumn{2}{c}{\begin{tabular}[c]{@{}c@{}}Question\\ Answering\end{tabular}}  \\ \cline{2-7}
                        & Average & $\Delta~(\uparrow)$
                        & Average & $\Delta~(\uparrow)$
                        & Average & $\Delta~(\uparrow)$ \\ \hline \hline
\textbf{Vicuna}
&       &                            
&       &                 
&       &                   \\
\hspace{0.7em}Zero-shot 
& 64.41 & 0.00 
& 75.00 & 0.00  
& 51.62 & 0.00      \\ \hline \hline
\colorbox{blue!10}{\textit{\textbf{Last}}}   
& & & & & &           \\
\hspace{0.5em} LwF \cite{li2017learning}
& 70.78 & +6.37
& 70.46 & -4.54
& 57.18 & +5.56 \\
\hspace{0.5em} EWC \cite{kirkpatrick2017overcoming}
& 47.57 & -16.84
& 72.37 & -2.63
& 52.42 & +0.80 \\
\hspace{0.5em} GMM \cite{cao2024generative}
& 60.62 & -3.79
& 73.83 & -1.17
& 54.93 & +3.31 \\
\hspace{0.5em} EProj \cite{he2023continual}
& 59.26 & -5.15
& 71.41 & -3.59
& 58.97 & +7.35  \\
\hspace{0.5em} MoEAdapter \cite{yu2024boosting} 
& 66.19 & +1.78
& 75.53 & +0.53
& 47.60 & -4.02 \\
\rowcolor[gray]{0.93}
\hspace{0.5em} MVP (Ours) 
& \textbf{86.68} & \textbf{+22.27}
& \textbf{77.55} & \textbf{+2.55}
& \textbf{70.93} & \textbf{+19.31} \\
 \hline \hline
\colorbox{red!10}{\textit{\textbf{Avg}}}       
& & & & & &           \\
\hspace{0.5em} LwF \cite{li2017learning}
& 62.88 & -1.53
& 65.36 & -9.64
& 52.72 & +1.10 \\
\hspace{0.5em} EWC \cite{kirkpatrick2017overcoming}
& 47.43 & -16.98 
& 63.03 & -11.97
& 40.43 & -11.19 \\
\hspace{0.5em} GMM \cite{cao2024generative}
& 46.76 & -17.65
& 72.54 & -2.46
& 45.89 & -5.73 \\
\hspace{0.5em} EProj \cite{he2023continual}
& 47.24 & -17.17
& 72.41 & -2.59
& 45.54 & -6.08 \\
\hspace{0.5em} MoEAdapter \cite{yu2024boosting} 
& 67.84 & +3.43
& 75.53 & +0.53
& 51.91 & +0.29 \\
\rowcolor[gray]{0.93} 
\hspace{0.5em} MVP (Ours)
& \textbf{81.67} & \textbf{+17.26}
& \textbf{77.12} & \textbf{+2.12}
& \textbf{60.87} & \textbf{+9.25}       \\ \hline \hline
\colorbox{green!10}{\textit{\textbf{Transfer}}}
&  &  
&  &  
&  &            \\
\hspace{0.5em} LwF \cite{li2017learning}
& 70.70 & +6.29
& 63.21 & -11.79
& 49.87 & -1.75 \\
\hspace{0.5em} EWC \cite{kirkpatrick2017overcoming}
& 57.77 & -6.64
& 61.92 & -13.08
& 39.85 & -11.77 \\
\hspace{0.5em} GMM \cite{cao2024generative}
& 55.01 & -9.40
& 72.43 & -2.57
& 44.74 & -6.88 \\
\hspace{0.5em} EProj \cite{he2023continual}
& 50.81 & -13.60 
& 72.67 & -2.33 
& 43.72 & -7.90           \\
\hspace{0.5em} MoEAdapter \cite{yu2024boosting} 
& 69.07 & +4.66
& 75.92 & +0.92
& 49.86 & -1.76 \\
\rowcolor[gray]{0.93} 
\hspace{0.5em} MVP (Ours)
& \textbf{76.09} & \textbf{+11.68}
& \textbf{77.13} & \textbf{+2.13}
& \textbf{55.98} & \textbf{+4.36}    \\ \toprule
\end{tabular}%
}
\end{table}

\begin{figure*}[t!]
    \centering
    \includegraphics[width=0.94\textwidth]{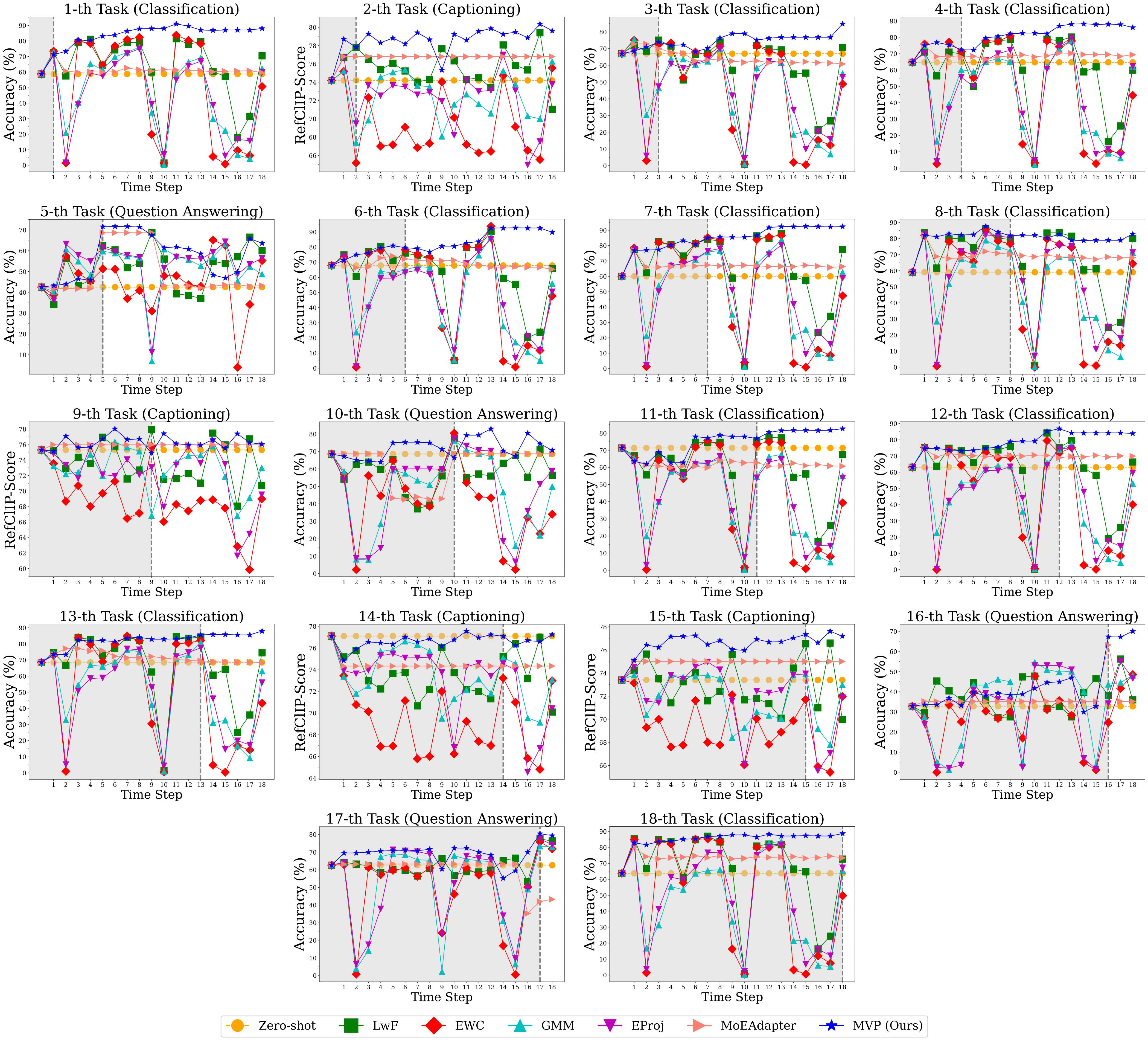}
    \caption{Results of training with a random task order, showing the performance of each task at each time step.
    The yellow horizontal line represents the zero-shot performance of the pre-trained
    model. 
    The gray area represents the time steps preceding training for specific tasks, and the vertical line indicates the time step at which the corresponding task is learned.
    Best viewed in color.}
    \label{fig: another results} 
\end{figure*}

\section{Additional Experimental Results}
\noindent{\textbf{Results on a Different Task Order.}}
The task order used in Tables \ref{tab:main table} and \ref{tab:another VLM} of the main paper follows the sequence of ten classification tasks, four captioning tasks and four question answering tasks.
Since captioning and question answering tasks do not have explicit classes, experiments on different class orders are omitted.
To evaluate the robustness of the proposed method to a different task order, we conducted additional experiments.
We randomly shuffled the task order used in Table \ref{tab:main table} of the main paper while keeping all other implementation details the same.
Table \ref{tab:different task order} reports the experimental results of learning with a random task order.
Overall, the proposed method outperforms other continual learning methods.
Most competitors show unsatisfactory performance for the three metrics compared to Zero-Shot.
Additionally, we report the task-specific performance measured at each time step in Figure \ref{fig: another results}.
While other continual learning methods exhibit large performance variations depending on the type of tasks being learned, the proposed method demonstrates robustness with respect to the task order.

\renewcommand{\arraystretch}{1.0}
\begin{table*}[t]
\caption{Language instructions for each type of task.}
\footnotesize
\label{tab: details of instruction}
\begin{tabularx}{\textwidth}{c|X}
\bottomrule
Task & \textbf{Classification} \\ \hline 
Default   & $\cdot$ What is this photo of?               \\ \hline
Generated & \begin{tabular}[t]{@{}p{\linewidth}@{}}
$\cdot$ Analyze the image to identify its most detailed category.\\ 
$\cdot$ The most detailed category should be determined based on the image's visual features.\\ 
$\cdot$ Assign the image to the most detailed category by examining its characteristics.\\ 
$\cdot$ Determine which detailed category best fits the given image and identify it.\\
$\cdot$ Based on the primary subject, identify the detailed category of the image.\\
$\cdot$ The image should be assessed to determine the most detailed category.\\
$\cdot$ Identify the correct detailed category for the given image.\\
$\cdot$ Assign the image to its respective detailed category after carefully identifying its features.\\
$\cdot$ Give the image a detailed category by examining its characteristics.\\
$\cdot$ Given the image's appearance, identify the appropriate detailed category.
\end{tabular} \\ \hline \hline 
Task  & \textbf{Captioning} \\ \hline     
Default   & $\cdot$ Describe the given image with a short sentence \\ \hline
Generated & \begin{tabular}[t]{@{}p{\linewidth}@{}}
$\cdot$ Describe the scene in detail by capturing its key visual elements, interactions, and overall atmosphere.\\ 
$\cdot$ The scene contains various elements, and it is essential to describe them clearly while maintaining a concise yet informative structure.\\
$\cdot$ Identify and describe the key subjects, their actions, and the interactions that define the scene comprehensively.\\
$\cdot$ A meaningful explanation should effectively describe the scene by considering its composition, significant events, and contextual details.\\
$\cdot$ The scene consists of numerous visual details, so it is important to describe them with clarity to ensure a precise understanding.\\
$\cdot$ Observe the scene carefully and describe the relationships between objects, people, and their interactions within the given context.\\
$\cdot$ Given the scene in the image, provide a well-structured statement that naturally describes its key features and dynamics.\\ 
$\cdot$ A well-formed caption must thoroughly describe the significant aspects of the scene, ensuring an accurate depiction of its core elements. \\
$\cdot$ As the scene unfolds dynamically, it is necessary to describe its core features with well-structured and concise sentences.\\
$\cdot$ Carefully examine the objects and contextual details in the scene to naturally describe its composition, interactions, and overall atmosphere. 
\end{tabular} \\ \hline \hline
Task  & \textbf{Question Answering} \\ \hline  
Default & \{Sample-wise Question\} \\ \hline
Generated & N/A \\ \toprule
\end{tabularx}
\end{table*}
\renewcommand{\arraystretch}{1.0}

\renewcommand{\arraystretch}{0.9}
\begin{table}[t]
\centering
\caption{
Results using generated instructions.}
\label{tab:other instructions}
\huge
\resizebox{\columnwidth}{!}{%
\begin{tabular}{l|cc|cc|cc}
\bottomrule
\multirow{2}{*}{Method} \hspace{1em} & \multicolumn{2}{c|}{Classification} & \multicolumn{2}{c|}{Captioning}   & \multicolumn{2}{c}{\begin{tabular}[c]{@{}c@{}}Question\\ Answering\end{tabular}}  \\ \cline{2-7}
                        & Average & $\Delta~(\uparrow)$
                        & Average & $\Delta~(\uparrow)$
                        & Average & $\Delta~(\uparrow)$ \\ \hline \hline
\textbf{Vicuna}
&       &                            
&       &                 
&       &                   \\
\hspace{0.5em}Zero-shot 
& 64.41 & 0.00 
& 75.00 & 0.00  
& 51.62 & 0.00      \\ \hline \hline
\colorbox{blue!10}{\textit{\textbf{Last}}}       
& & & & & &           \\
\hspace{0.5em}MVP w/ Default
& 85.87 & +21.46
& 77.75 & +2.75
& 76.34 & +24.72           \\
\hspace{0.5em}MVP w/ Generated
& 84.99 & +20.58
& 78.25 & +3.25
& 75.77 & +24.15   \\  
 \hline \hline
\colorbox{red!10}{\textit{\textbf{Avg}}}    
& & & & & &           \\
\hspace{0.5em}MVP w/ Default
& 83.28 & +18.87
& 76.94 & +1.94
& 57.68 & +6.06       \\ 
\hspace{0.5em}MVP w/ Generated
& 84.19 & +19.78
& 74.44 & -0.56
& 54.28 & +2.66          \\ \toprule \bottomrule
\colorbox{green!10}{\textbf{\textit{Transfer}}}
&  &  
&  &  
&  &            \\
\hspace{0.5em}MVP w/ Default
& 78.45 & +14.04
& 76.16 & +1.16
& 50.89 & -0.73    \\   
\hspace{0.5em}MVP w/ Generated
& 79.38 & +14.97
& 72.64 & -2.36
& 45.51 & -6.11        \\\toprule
\end{tabular}%
}
\renewcommand{\arraystretch}{1.0}
\end{table}

\noindent{\textbf{Analysis on Language Instructions.}}
Additionally, to analyze the robustness of the proposed instruction-grounded visual projector when multiple textual instructions are provided per task, we conducted an additional experiment.
We generated ten language instructions using GPT-4o \cite{achiam2023gpt} for image classification and captioning tasks.
The generated instructions for each task category are shown in the generated response in Table \ref{tab: details of instruction}.
For tasks belonging to the corresponding category, one of the ten generated instructions was randomly selected for each sample during training.
The experimental results are reported in Table \ref{tab:other instructions}.
The results indicate that the proposed method exhibits robustness, as its performance remains stable regardless of variations in instructions, provided that the instructions maintain contextual consistency.
Although an increase in the number of instructions leads to a slight decline in the transfer performance of the captioning and question answering tasks, the performance measured in \textit{Last} is comparable to those achieved using the default instructions.

\noindent{\textbf{Analysis on Computational Cost.}}
We analyzed the computational cost of MVP by measuring the wall-clock time and VRAM usage for the experiments in Table \ref{tab:main table} of the main paper.
The results are summarized in Table \ref{tab: efficiency_comparison}.
During its main training phase, MVP requires 20 hours and 48 minutes, a negligible time overhead (1.08$\times$) compared to GMM. 
This efficiency stems from freezing the computationally expensive vision and language backbones of the VLM while updating only the lightweight components: the router, the set of expert visual projectors, and the learnable pruning vector $E^t$.
In terms of memory, MVP requires 23.9 GB of VRAM, a marginal increase (1.16$\times$) over GMM from the use of multiple experts.
Notably, the subsequent prune and finetune stages are highly efficient, each completing in under 25 minutes and requiring less than 4.0 GB of VRAM. 
This low cost demonstrates that subsequent learning phase within our framework is highly practical.

\renewcommand{\arraystretch}{1.0}
\begin{table}[t!]
\caption{Analysis of computational cost.}
\label{tab: efficiency_comparison}
\centering
\huge
\resizebox{\columnwidth}{!}{%
\begin{tabular}{l|ccc}
\hline
\textbf{Vicuna}   & GMM (train) & EProj (train) & MVP (train / prune / finetune) \\ \hline
Wall-clock time   & 19h 58min  & 20h 02min & 20h 48min / 24min / 22min \\  
Memory (VRAM)     & 20.5GB     & 20.6GB   & 23.9GB    / 4.0GB / 3.1GB \\  \hline
\end{tabular}%
}
\renewcommand{\arraystretch}{1.0}
\end{table}

\renewcommand{\arraystretch}{1.0}
\begin{table}[t!]
\centering
\caption{Results on SEED-Bench after training all tasks.}
\tiny
\label{tab:seed_bench_results}
\large
\resizebox{\columnwidth}{!}{%
\begin{tabular}{l|ccccc}
\hline
\textbf{LLaMa-2}             & Zero-Shot & GMM   & EProj & MoEAdapter & MVP  \\ \hline
SEED-Bench \cite{li2023seed} & 42.66     & 32.08 & 29.69 & 34.72      & 42.59   \\ 
\hline
\end{tabular}%
}
\end{table}

\noindent{\textbf{Results on Standard VLM Benchmark.}}
To measure generalization capability, we evaluated the performance on SEED-Bench \cite{li2023seed} after their training on all tasks.
As shown in Table \ref{tab:seed_bench_results}, the compared methods exhibit a significant performance degradation from the Zero-Shot score, with a performance gap ranging from 7.94 to 12.97.
In contrast, MVP achieves a score of 42.59, which is the closest to the Zero-Shot performance. 
This performance retention is attributed to our adaptive knowledge aggregation strategy.
For unseen data, this strategy minimizes the influence of the trained experts and relies on the retained knowledge of the original pre-trained projector, thus preventing performance degradation.

\end{document}